\documentclass[letterpaper, 10 pt, conference]{ieeeconf}  

\IEEEoverridecommandlockouts                              
\usepackage[utf8]{inputenc}
\usepackage[T1]{fontenc}

\usepackage[pdftex]{graphicx}
\usepackage{adjustbox}
\usepackage{multirow}
\usepackage{mathptmx} 
\usepackage{amsmath} 
\usepackage{amssymb} 
\usepackage{subcaption}
\usepackage{xcolor}
\usepackage[ruled,vlined,linesnumbered]{algorithm2e}
\usepackage{balance}
\usepackage{microtype}
\newcommand{\boldparagraph}[1]{\vspace{0.2cm}\noindent{\bf #1:} }

\usepackage{placeins}
\usepackage{float}

\definecolor{red}{rgb}{1.0,0.0,0.0}
\definecolor{orange}{rgb}{1.0,0.65,0.0}
\definecolor{blue}{rgb}{0.0, 0.0, 1.0}
\definecolor{green}{rgb}{0.0, 1.0, 0.0}

\title{\LARGE \bf
Frustum Fusion: Pseudo-LiDAR and LiDAR Fusion for 3D Detection
}

\author{Farzin Negahbani$^{1,2}$, Onur Berk Töre$^{1,2}$, Fatma Güney$^{1,3}$ and Baris Akgun$^{1,2}$

\thanks{$^{1}$KUIS AI Lab, Koc University}%
\thanks{$^{2}$Autonomous, Learning and Interactive Agents Research Group}%
\thanks{$^{3}$Autonomous Vision Research Group}%
}

\begin{document}

\maketitle
\thispagestyle{empty}
\pagestyle{empty}

\begin{abstract}
Most autonomous vehicles are equipped with LiDAR sensors and stereo cameras. The former is very accurate but generates sparse data, whereas the latter is dense, has rich texture and color information but difficult to extract robust 3D representations from. In this paper, we propose a novel data fusion algorithm to combine accurate point clouds with dense but less accurate point clouds obtained from stereo pairs. We develop a framework to integrate this algorithm into various 3D object detection methods. Our framework starts with 2D detections from both of the RGB images, calculates frustums and their intersection, creates Pseudo-LiDAR data from the stereo images, and fills in the parts of the intersection region where the LiDAR data is lacking with the dense Pseudo-LiDAR points. We train multiple 3D object detection methods and show that our fusion strategy consistently improves the performance of detectors.
\end{abstract}

\section{INTRODUCTION}
    	


The advent of autonomous vehicles comes with many promises as well as challenges. The vehicles need to understand their environment to make decisions. One related challenge is detecting other agents, such as other cars, bicycles and pedestrians, that affect these decisions. These vehicles commonly employ both RGB cameras and LiDAR sensors to perceive their environment. As such, there exist approaches that make use of these modalities, either individually or combined, for detection. 

There is a notable performance gap between point-based methods and image-based methods for 3D object detection. Albeit including rich and dense texture and color information, extracting 3D information from images, even from stereo pairs, is challenging and error-prone. 
The learned representations from images hold promise but their performance still falls behind methods using LiDAR data which is far more accurate and reliable. On the other hand, LiDAR data is sparse compared to images. We argue that RGB images and LiDAR data are complementary in nature, and efficiently combining them will increase the performance of 3D detection. 

\begin{figure}[t]
            \centering
            \includegraphics[width=\columnwidth]{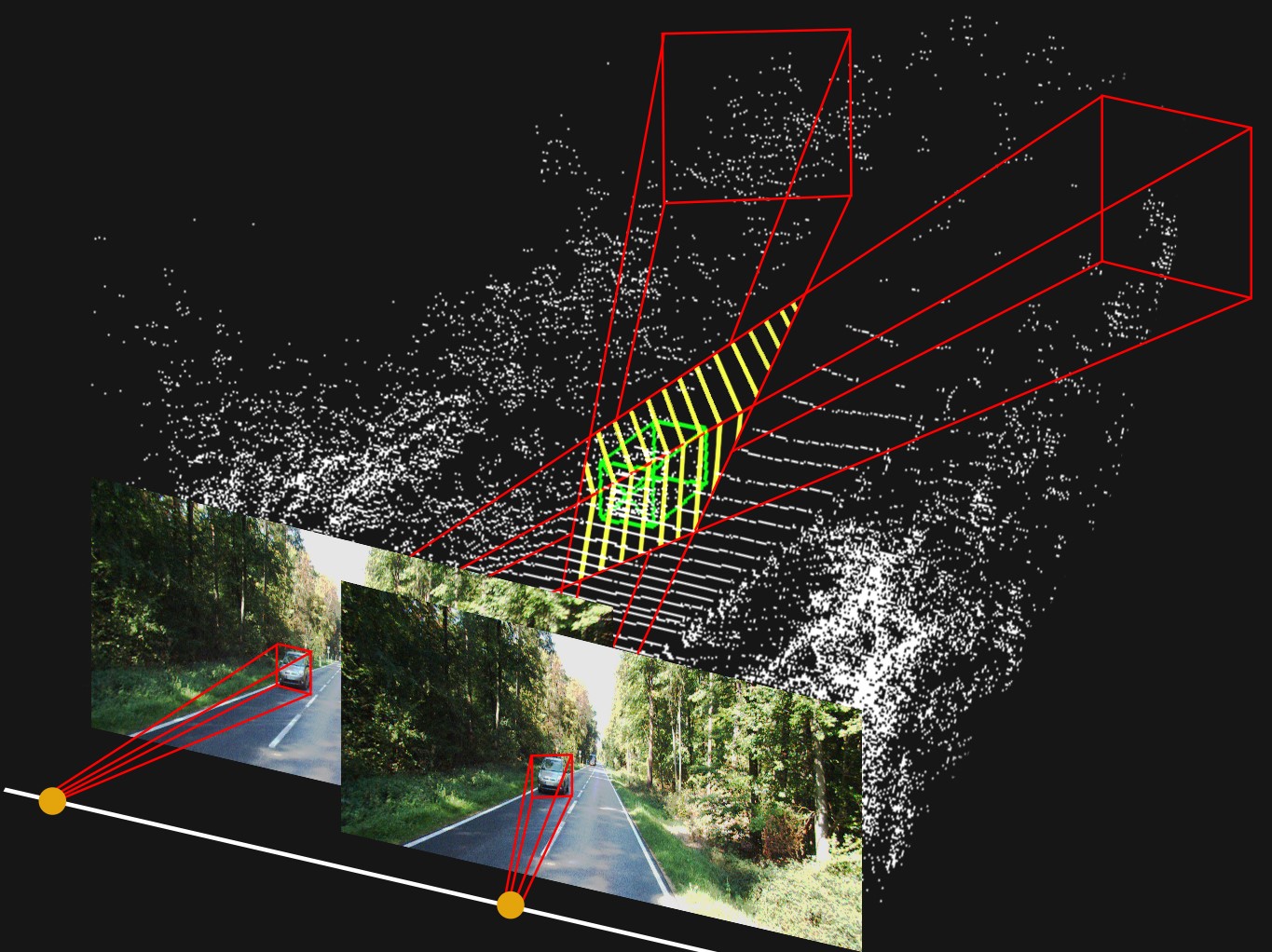}
            \caption{{\bf Frustum Fusion using Two Frustums.} The stripped region shows the intersection
            volume of the two frustums induced by the 2D detections on stereo images.}
            \label{fig:visual_concept}
    \end{figure}

In this paper, we propose a framework that combines 3D information extracted from stereo image data with LiDAR data 
and introduce a novel yet simple point cloud fusion algorithm. We first detect the object in the RGB images. In parallel, we calculate a dense 3D representation (Pseudo-LiDAR++ (PL++), \cite{you2019pseudoLiDAR}) from the stereo image pair. We then calculate the frustums induced in the 3D space by the 2D detections and find the LiDAR and PL points that fall in their intersection. We find the PL points where the LiDAR data is lacking within the intersection and combine them with the LiDAR data. Lastly, any point-based approach can be used to perform 3D detection on this fused data.
Visual depiction of these steps can be seen in Fig.~\ref{fig:visual_concept}. We evaluate our framework on the KITTI 3D object detection dataset \cite{Geiger2012CVPR} and show that it boosts the performance of multiple point-based 3D detectors. Our contributions can be summarized as follows: 
\begin{enumerate}
    \item A framework to combine RGB stereo images and LiDAR data
    \item A point cloud fusion algorithm 
    to efficiently combine accurate but sparse 
    LiDAR data with dense point cloud data obtained with the PL++ approach. 
\end{enumerate}

\section{Related Work}
    Learning-based 3D object detection methods can be categorized by the input representation they use: 2D RGB images, LiDAR/point clouds, and a combination of the two.
    While convolutional networks show a remarkable performance in detecting objects on 2D images, there is a considerable gap in 3D object detection accuracy between image-based methods and point-based ones.
    In this section, we follow an input-based organization with a focus on the methods that use a combination of the two.

    \begin{figure*}[!thp]
        \centering
        \includegraphics[width=0.99\textwidth]{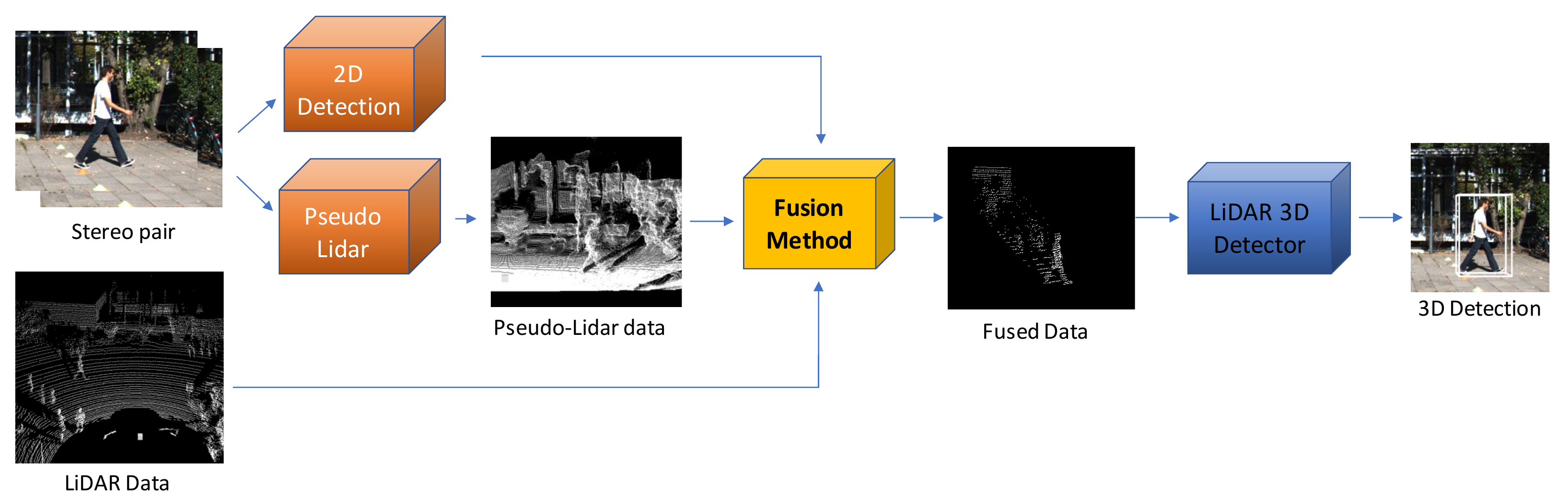}
        \caption{{\bf Overall Framework.} The process starts with 2D detections and Pseudo-LiDAR (PL) estimation from the stereo pair. The input to our fusion algorithm are the PL and LiDAR points at the intersection of two frustums induced by the 2D detections. In the fusion step, we combine sparse but accurate LiDAR points with dense but not as accurate PL points by using the distance of PL points to LiDAR as an indication for accuracy. Lastly, we perform 3D detection by using a point-based method 
        on the fused data. }
        \label{fig:pipeline}
    \end{figure*}

    
    \boldparagraph{RGB Images} Stereo RCNN \cite{li2019stereo} builds on top of a successful 2D detector \cite{faster-rcnn} to perform 3D detection using stereo images.
    Inspired by the representational power of point-based methods, Pseudo-LiDAR \cite{wang2018pseudoLiDAR} generates LiDAR-like data 
    by first performing depth estimation from stereo images.
    Since the success of this method depends on the accuracy of depth estimation, Pseudo-LiDAR++ \cite{you2019pseudoLiDAR} improves detection accuracy by applying a graph-based depth correction. 
    In order to reduce the variance of depth estimation, \cite{li2020confidence} propose to decode foreground and background separately by incorporating confidence values into the pipeline.
    DSGN \cite{chen2020dsgn} builds a 3D geometric volume to encode the geometric information from 2D features and learns depth and semantic cues from the built volume.

    
    
    There is an unavoidable information loss while extracting 3D information from images even in case of a stereo setup. More advanced methods improve depth estimation from images but there is still a considerable perfromance gap with methods that use LiDAR data. In this paper, we utilize image-based depth estimation in object regions where the sparse LiDAR data is missing.

    \boldparagraph{Point Clouds} Sparsity makes point-clouds a challeging input for learning. 
    PointNet \cite{qi2016pointnet} is the first learning-based method that work on raw point cloud data by exploiting spatial independence of points.
    Qi et al.\cite{qi2018frustum} first leverages 2D detections for finding frustums of candidate points and then learns to detect objects in the frustum using PointNet and  PointNet++ \cite{qi2017pointnet++}. 
    Lang et al.\cite{lang2019pointpillars} employ PointNet within an encoder that forms points in pillars (i.e. column format) and further applies a 2D convolution layer.  
    Instead of performing a single-stage as in the previous methods, PointRCNN \cite{shi2019pointrcnn} and Fast PointRCNN \cite{chen2019fast} use a second stage to refine 3D detections at the cost of computational complexity.


    \boldparagraph{Voxel Space} Another group of approaches convert point cloud data into a voxelized space for efficiency and abstraction.
    \cite{zhou2018voxelnet} learn 3D convolutions in voxel space to perform amodal bounding box estimation and object classification. Instead of costly dense 3D convolutions, \cite{yan2018second} further improve efficiency by using sparse 3D convolutions to exploit the sparse nature of point cloud data.
    PV-RCNN \cite{shi2020pv} incorporates PointNet features into voxel-based abstraction to improve 3D detection performance.

    Our framework can be used with both direct point cloud-based methods or voxel space methods and increase their performance.
    
    \boldparagraph{Data Fusion} Data fusion approaches aim to improve performance by combining information from different sources.
    Early studies MV3D\cite{chen2017multi}, AVOD \cite{ku2018joint}, and \cite{yan2018second} project point clouds into the bird-eye-view (BEV) and fuse extracted features from BEV and 2D images to perform 3D object detection. However, this kind of fusion approaches cannot reach the performance of the point-based methods. Although 2D RGB images provide rich semantic information which is missing in LiDAR, fusing dense image data with sparse point cloud data is challenging. Pang et al. \cite{pang2020clocs} utilize 2D and 3D detection candidates by first aligning them into a joint representational space and then performing learning in this space. This is one of the best performing fusion methods due to joint exploration of geometric and semantic features in a large space of 2D-3D detection pairs before applying Non-maximum Suppression (NMS) algorithm.
    Compared to these fusion methods, we leverage the 2D bounding boxes to narrow down the search space on the point cloud to the intersection of two frustums, and fuse Pseudo-LiDAR data extracted from 2D images to enrich the data. 

\section{Methodology}
    
Our proposed framework consists of four main stages; (1) 2D object detection on each of the stereo RGB images to extract bounding box and class information, (2) generating Pseudo-LiDAR data from a pair of stereo images using the PL++ method \cite{you2019pseudoLiDAR}, (3) calculating the intersection of the frustums induced by 2D detections and fusing the LiDAR and the Pseudode-LiDAR data in this region, and (4) performing 3D object detection on fused point cloud data.
Fig.~\ref{fig:pipeline} shows the overall framework together with interactions between different stages.

    We start by detecting objects on both stereo images. In order to eliminate the effect of a 2D object detector and show the power of the proposed fusion method,
    we use ground truth detections in this paper. 
    Next, we explain the generation of Pseudo-LiDAR point cloud data from stereo images and our early fusion strategy.
    
    
    \subsection{Pseudo-LiDAR}
    In the second stage, we generate additional point cloud data from stereo images. The LiDAR data can be sparse due to properties of sensor or missing measurements. This sparsity can cause problems when detecting small objects which are often represented only with a few points. Additional points can be incorporated based on 
    noisy, dense depth estimation from stereo images. The idea of generating stereo depth estimations similar to LiDAR data has first been explored in Pseudo-LiDAR \cite{wang2018pseudoLiDAR} and then extended in Pseudo-LiDAR++ \cite{you2019pseudoLiDAR}. In this paper, we follow a similar approach to Pseudo-LiDAR++ for generating a point cloud set from stereo images and then incorporate it into LiDAR data with our fusion algorithm.
    
    
    By following the representation proposed in Pseudo-LiDAR, we calculate the horizontal, vertical, and depth coordinates, $(x, y, z)$, of a point corresponding to pixel $(u,v)$ on the disparity map $\mathbf{Y}$ as follows:
    \begin{equation}
        z = \dfrac{f_u \;\times\;  b }{{\mathbf{Y}}(u,v)}
        \label{eq:depth}
    \end{equation}  
    \begin{equation}
        x = \dfrac{(u \; - \; c_u)\;\times \;z }{f_u},
        \label{eq:width}
    \end{equation}  
    \begin{equation}
        y = \dfrac{(v \; - \; c_v)\;\times \; z }{f_v}
        \label{eq:height}
    \end{equation}  
    %
    where $f_u$ is the horizontal focal length, $f_v$ is the vertical focal length, $b$ is the baseline corresponding to horizontal offset between the cameras, and the $c_u$ and $c_v$ are camera center in pixels.
    Following the improvements proposed in Pseudo-LiDAR++ \cite{you2019pseudoLiDAR}, we directly optimize a depth loss using a cost volume in order to learn parameters adjusted according to depth values.
    
    In Pseudo-LiDAR++, estimated dense depth map is sampled to generate a sparse set of points which resemble typical
    LiDAR data (e.g. a 64-beam Velodyne). 
    This random sampling may omit some points that are important for detecting objects. Therefore, we use the dense points as a starting point for our fusion approach.
    
    \SetKwInput{KwInput}{Input}
    \SetKwInput{KwOutput}{Output}

    \begin{algorithm}[tb]
    \small
    \SetAlgoLined
        \KwInput{Pseudo-LiDAR Data (PLD)  \& LiDAR Data (LD) \& 2D Bounding Boxes (BBs)}
        \KwResult{Fused Set}
        
        $lidar$ $\gets$ ext{$\_$}intersection(LD, 2D BBs) \\
        $pseudolidar$ $\gets$ ext{$\_$}intersection(PLD, 2D BBs) \\
        $fused{\_}set \gets lidar$ \\
        \For{$point$ in $pseudolidar$} {
           $min{\_}dist$ $\gets$ find{$\_$}min{$\_$}euclidean{$\_$}dist(point, lidar) 
           
           \If{$min{\_}dist \geq \tau$}{
                $fused{\_}set$.append(point)
            }
         }
         
        \Return $fused\_set$
    \caption{Frustum Fusion Algorithm}
    \label{alg_1}
    \end{algorithm}
    
\begin{figure}[!thb]
\centering
\begin{subfigure}[b]{\columnwidth}
    \centering
    \includegraphics[width=0.86\columnwidth]{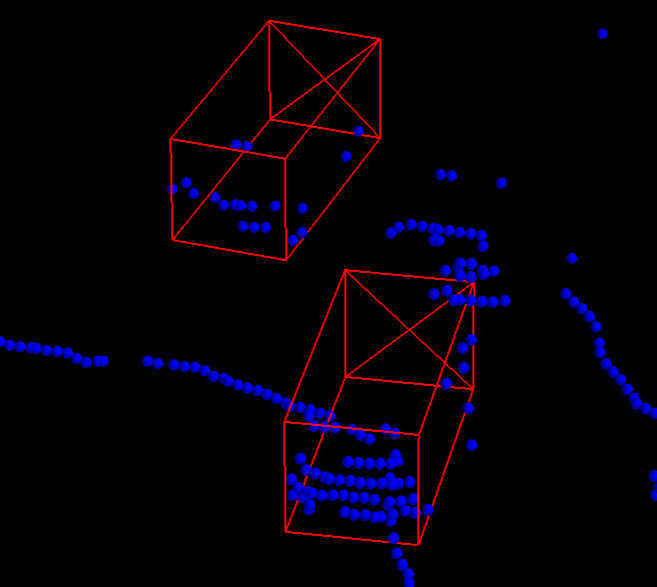}
    \caption{LiDAR Data}
    \label{fig:Velodyne}
\end{subfigure}
\begin{subfigure}[b]{\columnwidth}
    \centering
    \includegraphics[width=0.86\columnwidth]{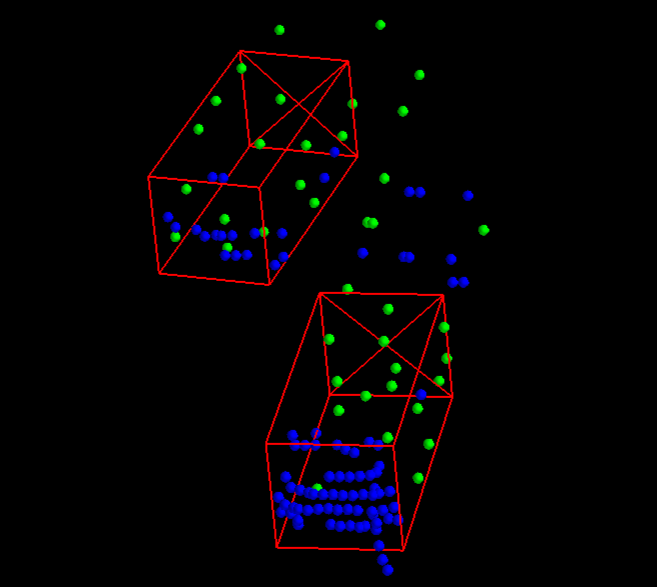}
    \caption{Fused Data}
    \label{fig:fused}
\end{subfigure}
\caption{{\bf Velodyne vs. Fused Data.} This figure compares the LiDAR data from the Velodyne sensor (a) to the result of fusion (b) as the combination of points extracted from the intersection of frustums and the Pseudo-LiDAR points that are selected by our algorithm. Our fusion strategy removes unnecessary points outside the detection boxes and adds points from PL to fill in areas with few LiDAR points, resulting in densely populated areas where objects exist. Points  in {\color{green}green} merged from PL and  in {\color{blue}blue} are LiDAR data.
}
 \label{fig:Velodyne_vs_fused}
 \vspace{-32pt}
\end{figure}

\subsection{Fusion Method}
At this stage, we have two sources of data: accurate but sparse LiDAR data and dense but not as accurate Pseudo-LiDAR (PL) data.
Due to these differences, combining dense PL points directly with the LiDAR data reduces the overall accuracy. 
Furthermore, typical LiDAR data is sparse and the existing 3D object detectors are designed by taking this sparsity into account.
Thus, our fusion algorithm finds the PL points that enrich our data instead of overwhelming it. It works by finding the PL points that are within the frustum intersection but away from existing LiDAR points. This is opposed to random samplin usually employed by existing methods.
The pseudo-code of our fusion algorithm is given in Alg.~\ref{alg_1}.

Before the fusion step, we find the candidate points that may correspond to the object in the point cloud space, represented by the ext{\_}intersection (lines 1-2) function in Alg.~\ref{alg_1}. The goal of this is to reduce the costly distance calculations during fusion and to reduce the number of irrelevant points (e.g. instead of the background or another object) for 3D detection. We utilize the 2D detections from the stereo images towards this end. We first calculate the frustums induced by these detections on the point cloud space. We use the frustum extraction method used in \cite{qi2018frustum} but for both images instead of one. Then we find the points in the point cloud data that are at the intersection of these frustums, both for the LiDAR data and the Pseudo-Lidar data. The name of our approach, Frustum Fusion, comes from this step. Fig.~\ref{fig:visual_concept} shows a visual depiction of the frustums from detections and their intersections.

At this point, we have the points that fall within the intersection of two frustums for both the LiDAR and the pseudo-LiDAR data. In order to decide which points should be fused (i.e. taken from the pseudo-LiDAR data), we use Euclidean distance between the points. If the distance between a pseudo-LiDAR point and the closest LiDAR point is higher than a given threshold ($\tau$), the PL point is added to the LiDAR data. This process is applied for all the PL points that are in the frustum intersection to generate the fused point cloud. An example LiDAR point cloud and the resulting fused data can be seen in Fig.~\ref{fig:Velodyne_vs_fused}. This figure shows the advatnages of taking the intersection of two frustums (removal of unnecessary points) and our distance based fusion algorithm (addition of relevant points with minimum extras). 

    \begin{table*}[ht]
        \caption{ \textbf{3D Object Detection Results.} We report $AP_{3D}$ (in \%) results on the KITTI validation set using 40 recall positions. The first row shows the baseline results, PV-RCNN\cite{shi2020pv} method
        , and the following rows belong to our fusion strategy using the same detector under different 
        thresholds for \textit{Easy}, \textit{Moderate}, and \textit{Hard} difficulty levels on KITTI validation set.
        }
        \label{PV-RCNN_table}
        \begin{adjustbox}{width=\textwidth,center}
            \centering
            \label{tab:my-table}
                \begin{tabular}{|c|c|c|c|l|c|c|c|l|c|c|c|l|l|}
                    \hline
                    \multirow{2}{*}{PV-RCNN \cite{shi2020pv}} & \multicolumn{4}{c|}{\textbf{\begin{tabular}[c]{@{}c@{}}Car $AP_{3D}$\end{tabular}}} & \multicolumn{4}{c|}{\textbf{\begin{tabular}[c]{@{}c@{}}Cyclist $AP_{3D}$\end{tabular}}} & \multicolumn{4}{c|}{\textbf{\begin{tabular}[c]{@{}c@{}}Pedestrian $AP_{3D}$\end{tabular}}} & \multicolumn{1}{c|}{\multirow{4}{*}{\textbf{\begin{tabular}[c]{@{}c@{}}all\\ tasks\\ mAP\end{tabular}}}} \\ \cline{2-13}
                    
                     & Easy & Moderate & Hard & \multirow{3}{*}{mAP} & Easy & Moderate & Hard & \multirow{3}{*}{mAP} & Easy & Moderate & Hard & \multirow{3}{*}{mAP} & \multicolumn{1}{c|}{} \\ \cline{1-4} \cline{6-8} \cline{10-12}
                     
                    \multirow{2}{*}{\textbf{Data}} & \multicolumn{3}{c|}{IoU} &  & \multicolumn{3}{c|}{IoU} &  & \multicolumn{3}{c|}{IoU} &  & \multicolumn{1}{c|}{} \\ \cline{2-4} \cline{6-8} \cline{10-12}
                    
                     & 0.7 & 0.7 & 0.7 &  & 0.5 & 0.5 & 0.5 &  & 0.5 & 0.5 & 0.5 &  & \multicolumn{1}{c|}{} \\ \hline
                    
                    LiDAR & 91.26 & 82.57 & 81.95 & 85.26 & 88.29 & 72.26 & 67.80 & 76.11 & 64.46 & 57.72 & 52.49 & 58.22 & 73.19 \\
                    FF $\tau$ = 0.25 & 92.94 & 86.55 & 83.89 & 87.79 & 89.94 & 68.88 & 63.97 & 74.26 & 64.90 & 57.87 & 52.90 & 58.56 & 73.53 \\
                    FF $\tau$ =  0.5 & 92.62 & 88.68 & 84.37 & 88.56 & 92.00 & 74.21 & 69.41 & 78.54 & \textbf{65.50} & \textbf{60.56} & 54.75 & \textbf{60.27} & 75.79 \\
                    FF $\tau$ =  0.6 & 93.30 & \textbf{89.04} & \textbf{86.59} & \multicolumn{1}{c|}{\textbf{89.64}} & 91.57 & 74.05 & 69.37 & \multicolumn{1}{c|}{78.33} & 65.15 & 59.99 & 55.31 & \multicolumn{1}{c|}{60.15} & \multicolumn{1}{c|}{76.04} \\
                    FF $\tau$ =  0.7 & 92.95 & 86.81 & 86.24 & 88.67 & 91.41 & 76.19 & 73.14 & 80.25 & 65.28 & 60.04 & \textbf{55.49} & \textbf{60.27} & \textbf{76.39} \\
                    FF $\tau$ =  0.9 & \textbf{93.45} & 86.76 & 84.14 & 88.12 & \textbf{92.99} & \textbf{78.75} & \textbf{73.57} & \textbf{81.77} & 62.14 & 57.59 & 53.25 & 57.66 & 75.85 \\ \hline
                    \end{tabular}
        \end{adjustbox}
    \end{table*}
    
\begin{table*}[ht]
    \centering
    \caption{\textbf{3D Detection Results According to Classes (Moderate).} We report the results of detecting Car (IoU=0.7), Cyclist (IoU=0.5), and Pedestrian (IoU=0.5) classes separately at the difficulty level moderate on the validation set. We compare the performance of different 3D detection methods using KITTI LiDAR only versus our fusion strategy with different thresholds in terms of $AP_{3D}$ / $AP_{BEV}$ (in \%). See the text for details.}
    \label{tab:moderate_summary}
    \begin{tabular}{c|c|c|c|c|c|c}
        \multicolumn{7}{c}{\textbf{Car Moderate $AP_{3D}$ / $AP_{BEV}$}} \\ \hline
        \multicolumn{1}{|c|}{\multirow{2}{*}{\textbf{Method}}} & \multicolumn{5}{c|}{\textbf{Data}} & \multicolumn{1}{|c|}{\multirow{2}{*}{\textbf{Max Improve}}} \\ \cline{2-6}
        \multicolumn{1}{|c|}{} & LiDAR & FF $\tau$ = 0.25 & FF $\tau$ = 0.5 & FF $\tau$ = 0.7 & FF $\tau$ = 0.9 &  \multicolumn{1}{|c|}{}  \\ \hline
        \multicolumn{1}{|c|}{PV-RCNN\cite{shi2020pv}} & 82.57 / 90.60 & 86.55 / 92.16 & 88.64 / 92.61 & \textbf{86.81} / \textbf{92.62} & 86.76 / 92.31 & \multicolumn{1}{|c|}{+6.07 / +2.02} \\ 
        \multicolumn{1}{|c|}{SECOND\cite{yan2018second}} & 81.08 / 89.27 & 84.62 / 88.86 & 84.28 / 90.83 & 84.54 / 91.03 & \textbf{84.75} / \textbf{91.32} & \multicolumn{1}{|c|}{+3.67 / +2.05} \\ 
        \multicolumn{1}{|c|}{PointRCNN\cite{shi2019pointrcnn}} & 80.23 /  88.81 & 83.19 / 90.02 & \textbf{84.10} / 90.51 & 83.87 / 90.04 & 84.03 / \textbf{90.56} & \multicolumn{1}{|c|}{+3.87 /  +1.75} \\ 
        \multicolumn{1}{|c|}{Part-A2Net\cite{shi2019points}} & 80.10 / 88.15 & 84.24 /  90.37 & \textbf{86.75} / \textbf{92.72} & 83.74 / 92.24 & 83.80 / 90.02 & \multicolumn{1}{|c|}{+6.65 / +4.57} \\ 
        \multicolumn{1}{|c|}{F-ConvNet\cite{wang2019frustum}} & 78.36 / 89.56 & 74.51 / 85.91 & 78.14 / 88.53 & \textbf{85.66} / 89.33 & 84.90 / 89.30 & \multicolumn{1}{|c|}{+7.30 / -0.23} \\ 
        \multicolumn{1}{|c|}{PointPillars\cite{lang2019pointpillars}} & 78.27 / 88.37 & 79.92 / 86.61 & 82.39 / 88.96 & \textbf{82.80} / \textbf{89.31} & 81.73 / 89.18 & \multicolumn{1}{|c|}{+4.53 / +0.94} \\ \hline
        \multicolumn{7}{c}{} \\
        \multicolumn{7}{c}{\textbf{Cyclist Moderate $AP_{3D}$ / $AP_{BEV}$}} \\ \hline
        \multicolumn{1}{|c|}{\multirow{2}{*}{\textbf{Method}}} & \multicolumn{5}{c|}{\textbf{Data}} & \multicolumn{1}{|c|}{\multirow{2}{*}{\textbf{Max Improve}}} \\ \cline{2-6}
        \multicolumn{1}{|c|}{} & LiDAR & FF $\tau$ = 0.25 & FF $\tau$ = 0.5 & FF $\tau$ = 0.7 & FF $\tau$ = 0.9 &  \multicolumn{1}{|c|}{}  \\ \hline
        \multicolumn{1}{|c|}{PV-RCNN\cite{shi2020pv}} & 72.26 / 75.59 & 68.88 / 71.32 & 74.21 / 77.92 & 76.19 / 79.20 & \textbf{78.75} / \textbf{81.16}& \multicolumn{1}{|c|}{+6.49 / +5.57} \\ 
        \multicolumn{1}{|c|}{SECOND\cite{yan2018second}} & 67.20 / 72.42 & 59.63 / 63.09 & 69.97 / 74.32 & \textbf{73.96} / \textbf{77.04} & 71.55 / 75.39 & \multicolumn{1}{|c|}{+6.76 / + 4.62} \\ 
        \multicolumn{1}{|c|}{PointRCNN\cite{shi2019pointrcnn}} & 73.13 / 75.37 & 69.61 / 70.58 & 77.67 / 82.27 & 75.75 / 79.39 & \textbf{81.96} / \textbf{84.80} & \multicolumn{1}{|c|}{+8.83 / +9.43} \\ 
        \multicolumn{1}{|c|}{Part-A2Net\cite{shi2019points}} & 72.75 /  76.87 & 71.92 / 73.49 & 78.33 / 79.64 & 79.04 / 81.94 & \textbf{79.27} / \textbf{82.29} & \multicolumn{1}{|c|}{+6.52 / +5.42} \\ 
        \multicolumn{1}{|c|}{F-ConvNet\cite{wang2019frustum}} & 77.99 / 87.16 & 67.21 / 75.41 & 78.43 / 87.64 & 85.87 / 87.15 & \textbf{88.24} / \textbf{89.51} & \multicolumn{1}{|c|}{+10.25 / +2.35} \\ 
        \multicolumn{1}{|c|}{PointPillars\cite{lang2019pointpillars}} & 63.45 /  68.16 & 52.96 / 55.53 & 56.33 / 60.96 & 66.02 / \textbf{71.58} &\textbf{66.60} / 70.89 & \multicolumn{1}{|c|}{+3.15 / +3.42} \\ \hline
        
        \multicolumn{7}{c}{} \\
        \multicolumn{7}{c}{\textbf{Pedestrian Moderate $AP_{3D}$ / $AP_{BEV}$}} \\ \hline
        \multicolumn{1}{|c|}{\multirow{2}{*}{\textbf{Method}}} & \multicolumn{5}{c|}{\textbf{Data}} & \multicolumn{1}{|c|}{\multirow{2}{*}{\textbf{Max Improve}}} \\ \cline{2-6}
        \multicolumn{1}{|c|}{} & LiDAR & FF $\tau$ = 0.25 & FF $\tau$ = 0.5 & FF $\tau$ = 0.7 & FF $\tau$ = 0.9 &  \multicolumn{1}{|c|}{}  \\ \hline
        \multicolumn{1}{|c|}{PV-RCNN\cite{shi2020pv}} & 57.72 / 61.18 & 57.87 / 62.96 & \textbf{60.56} / 65.18 & 60.04 / \textbf{65.78} & 57.59 / 63.51 & \multicolumn{1}{|c|}{+2.84 / +4.6} \\ 
        \multicolumn{1}{|c|}{SECOND\cite{yan2018second}} & 50.70 / 55.42 & \textbf{54.93} / \textbf{60.69} & 49.57 / 57.46 & 53.01 / 60.09 & 52.64 / 59.28 & \multicolumn{1}{|c|}{+4.23 / +5.27} \\ 
        \multicolumn{1}{|c|}{PointRCNN\cite{shi2019pointrcnn}} & 55.50 / 58.45 & 54.27 / 60.47 & 55.37 / 59.25 & \textbf{56.90} / \textbf{62.49} & 53.96 / 59.18 & \multicolumn{1}{|c|}{+1.40 / +4.04} \\ 
        \multicolumn{1}{|c|}{Part-A2Net\cite{shi2019points}} & 64.71 / 68.00 & 63.50 / 68.90 & \textbf{64.81} /\textbf{70.48} & 62.60 / 68.74 & 63.83 / 70.23 & \multicolumn{1}{|c|}{+0.10 / +2.48} \\ 
        \multicolumn{1}{|c|}{F-ConvNet\cite{wang2019frustum}} & 50.01 / 60.55 & 50.84 / 59.81 & 57.37 / 63.05 & \textbf{61.19} / \textbf{64.28} & 57.48 / 62.23 & \multicolumn{1}{|c|}{+11.18 / + 3.73} \\ 
        \multicolumn{1}{|c|}{PointPillars\cite{lang2019pointpillars}} & 48.13 / 53.67 & 51.78 / 56.73 & \textbf{54.44} / 61.35 & 49.60 / 56.02 & 53.99 / \textbf{61.54} & \multicolumn{1}{|c|}{+6.31 / +7.87} \\ \hline
        \end{tabular}
\end{table*}

After the fusion, the resulting point cloud can be input to
any 3D object detector that takes point clouds (e.g. from a LiDAR) as input. We experiment with different detection methods and show improvements for each due to our fusion strategy as explained next in Sect.~\ref{sect:detectors}. 

\section{Evaluation Setup} 
In this section, we first list the 3D detectors that we experiment with, then introduce the dataset and the metrics that we use for evaluation, and finally, we show and discuss our results.
\subsection{3D Detectors} \label{sect:detectors}
In order to evaluate the effect of the proposed fusion strategy, we experiment with several 3D object detectors and present results with and without our fusion strategy. In particular, we experiment with voxel-based (\cite{shi2020pv}), PointNet-based (\cite{shi2020pv, wang2019frustum,shi2019pointrcnn}), joint BEV and 2D feature-based (\cite{yan2018second,shi2019points}), and lastly with Pillar-based (\cite{lang2019pointpillars}) 3D detectors to evaluate our approach.
Our goal is to demonstrate that our fusion strategy can generalize over various input representations.
We compare the performance of the original detector to its performance within our framework by using different distance thresholds ($\tau$). 




\subsection{Dataset and Metrics}
We evaluate our approach on the "3D Object Detection Evaluation 2017" benchmark of the KITTI dataset \cite{Geiger2012CVPR}. This benchmark contains 7481 training and 7518 testing data points. The input to our method is the 
64-Beam Velodyne LiDAR sensor data and the RGB stereo pair. There are three classes of objects in the dataset; (1) Cars, (2) Cyclists, and (3) Pedestrians. The training set contains object annotations for the LiDAR point clouds and stereo images. We are interested in 3D amodal object detection ($AP_{3D}$) and 2D ground detection ($AP_{BEV}$) evaluations. The dataset classifies the difficulty of object detection based on the bounding box height, object occlusion and truncation percentages as \textit{Easy}, \textit{Moderate} and \textit{Hard}.

We further split the training set into train and validation sets by following the approach in \cite{NIPS2015_5644}. All methods are trained on this training set and tested on this validation 
set in terms of $AP_{BEV}$ and $AP_{3D}$ based on 40 recall points \cite{simonelli2019disentangling}. We use the default hyper-parameters of the methods for training.  We report results of the best-performing PV-RCNN\cite{shi2020pv} for all three difficulty levels while only reporting the moderate cases for the rest of the methods due to space concerns.




\section{Results and Discussion}
    
    
    

    
    
    We present our evaluation results in multiple tables. Table~\ref{PV-RCNN_table} shows the results of 3D detection using the PV-RCNN \cite{shi2020pv} method as our 3D detector for all difficulty levels of all tasks with multiple fusion thresholds. 
    Table~\ref{tab:moderate_summary} shows the results of 3D detection and birds-eye-view (BeV) detection of the moderate difficulty task with multiple 3D detectors and fusion thresholds for different object classes. We present our findings and discuss their implications next.
    
    \boldparagraph{Our fusion strategy increases the performance of all the 3D Detection methods for all three detection tasks and difficulty levels.} Our framework utilizing different thresholds outperform the LiDAR baseline in almost all object categories and difficulty levels for the PV-RCNN method as shown in Table~\ref{PV-RCNN_table}. The other three tables show the same results for the other methods, in both 3D and BeV detection for the moderate difficulty case. This implies that our method is well poised to improve the performance of different point-based 3D detectors with different characteristics.
    
    \boldparagraph{The best fusion thresholds may differ between classes and for a given method}. Table~\ref{PV-RCNN_table} shows that the best overall threshold is 0.7 meters. However, it is 0.6 for car, 0.9 for cyclist, and a tie between 0.5 and 0.7 for the pedestrian classes. Comparing the rows of other tables corresponding to different methods show similar results. Thresholds usually differ between classes. This is not a drawback for our method since the 2D detection part already predicts the class label and as a result, a specific threshold can be used down the pipeline. This result implies that we need to to utilize the object class for selecting the best fusion threshold. 
    
    \boldparagraph{The best fusion thresholds may differ between methods for a given class}. 
    The bold entries of Table~\ref{tab:moderate_summary} suggests that for the same class, different methods perform best at different fusion thresholds. A notable exception is the cyclist class which shows some consistency across different methods. This result implies that the fusion threshold should be tuned for a given 3D detection method.
    
    \boldparagraph{The best fusion thresholds are mostly the same for all difficulty cases of a given class.} Table~\ref{PV-RCNN_table} shows that the best threshold values between difficulty levels for a given class do not vary much. In case of a mismatch, the performance differences are small (e.g. 93.3 $AP_{3D}$ for $\tau=0.6$ vs 93.45 for $\tau=0.9$). This result implies that the selection of a single threshold does not result in a trade-off in terms of detection difficulty. However, the evidence for this only comes from the PV-RCNN method.
    

    
\begin{figure*}[t]
        \centering
        \includegraphics[width=\textwidth]{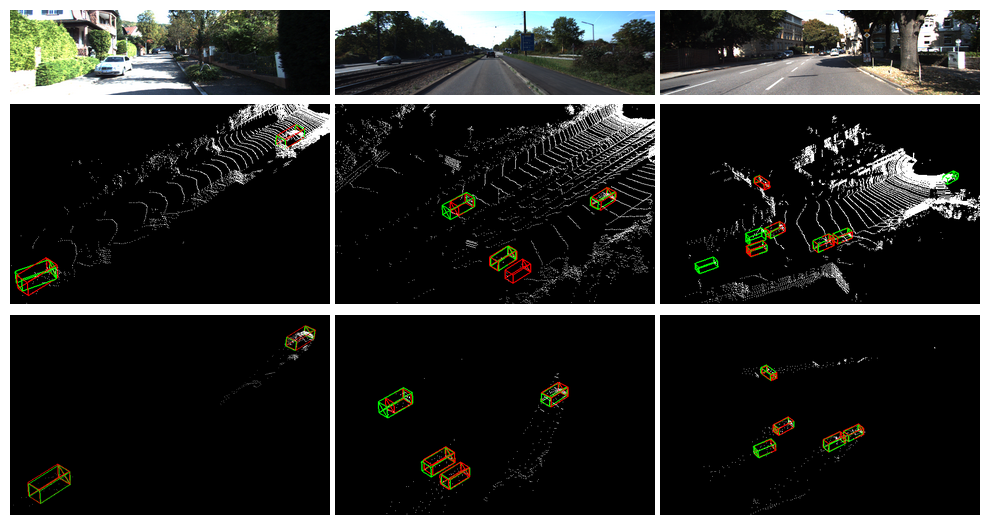}
        \caption{{\bf Qualitative Results.} The first row demonstrates RGB image of the scene. The second row shows the results of PV-RCNN \cite{shi2020pv} method using KITTI LiDAR data only. The third row shows the result of our fusion strategy using PV-RCNN as the 3D detector.
        Ground-truth boxes are shown in {\color{red}red} and predictions are in {\color{green}green}. Our fusion strategy reduces the number of false positives and results in predictions that better overlap with the ground-truth. (Best viewed in color)}
        \label{qual}
    \end{figure*}

    In addition to the quantitative results, we present qualitative results. The previous work \cite{wang2018pseudoLiDAR,you2019pseudoLiDAR} reports that the stereo depth estimation error becomes prohibitive for object detection beyond 30 meters on KITTI dataset. However, we observe that fusing PL data with the LiDAR data improves both close and distant objects.  Fig.~\ref{qual} depicts several examples comparing the results of the PV-RCNN method with KITTI LiDAR (the second row) and the fused data (the third row). 
    Note that adding the PL data helps to correctly estimate the 3D bounding box for the distant car in the most left sample.

\section{Conclusion}
The existing literature does not fully utilize stereo cameras and LiDAR sensors together which are readily available in autonomous vehicles. In this paper, we have proposed an algorithm, called Frustum Fusion, to combine sparse but accurate LiDAR data with dense but noisy Pseudo-LiDAR data obtained from stereo images. We have further introduced a framework to utilize our algorithm with six different 3D detection methods that take point clouds as input. 

We performed an extensive evaluation of these 3D detection methods for detecting different classes at different difficulty levels by changing fusion threshold. We showed that the methods trained with our fusion strategy outperform detectors trained with LiDAR data only for all object classes, difficulty levels, and 3D methods. Our results also show the need to tune the fusion threshold for a given method and a class, however, this is fixed during training and does not pose an issue during test time. 

In this paper, we opted to develop a data fusion algorithm, instead of another architecture. Our motivation was to utilize the full extent of the available sensors. As such, we argue that our contribution is vertical and can be used to boost the performance of future 3D detectors as well. 

Lastly, we propose some future directions to improve our work.
Despite limiting the candidate points to frustum intersections, the distance calculations for the fusion method can be costly. One could perform these computations in parallel or use approximate distance calculations for efficiency. 
Another promising direction in Pseudo-LiDAR is to utilize left-right and right-left estimations beyond consistency check for guiding fusion towards accurate PL points with high consistency.

\addtolength{\textheight}{-12cm}   

\balance


\bibliographystyle{unsrt}

\end{document}